\documentclass{article}
\usepackage{ijcai18}

\usepackage{times}
\usepackage{xcolor}
\usepackage{soul}
\usepackage[utf8]{inputenc}
\usepackage[small]{caption}
\usepackage{graphicx}
\usepackage{amsmath}

\title{Annotation-Free and One-Shot Learning for
Instance Segmentation of Homogeneous Object Clusters}

\author{
Zheng Wu$^1$, 
Ruiheng Chang$^1$, 
Jiaxu Ma$^1$,
Cewu Lu$^1$,
Chi-Keung Tang$^2$ 
\\ 
$^1$ Shanghai Jiao Tong University \\
$^2$ HKUST\\
\{14wuzheng,crh19970307,alexma,lucewu\}@sjtu.edu.cn,
cktang@cse.ust.hk
}

\begin{document}

\maketitle

\begin{abstract}
We propose a novel approach for instance segmentation given an image
of homogeneous object cluster (HOC).  Our learning approach is
one-shot because a single video of an object instance is captured
and it requires no human annotation.
Our intuition is that images of homogeneous objects can be effectively synthesized based on structure and illumination priors derived from real images. A novel solver is proposed that iteratively maximizes our structured likelihood to generate realistic images of HOC. Illumination transformation scheme is applied to make the real and synthetic images share the same illumination condition. Extensive experiments and comparisons are performed to verify our method. We build a dataset consisting of pixel-level annotated images of HOC. The dataset and code will be published with the paper.
\end{abstract}

\section{Introduction}
\begin{figure}
  \begin{center}
    \includegraphics[width=\linewidth]{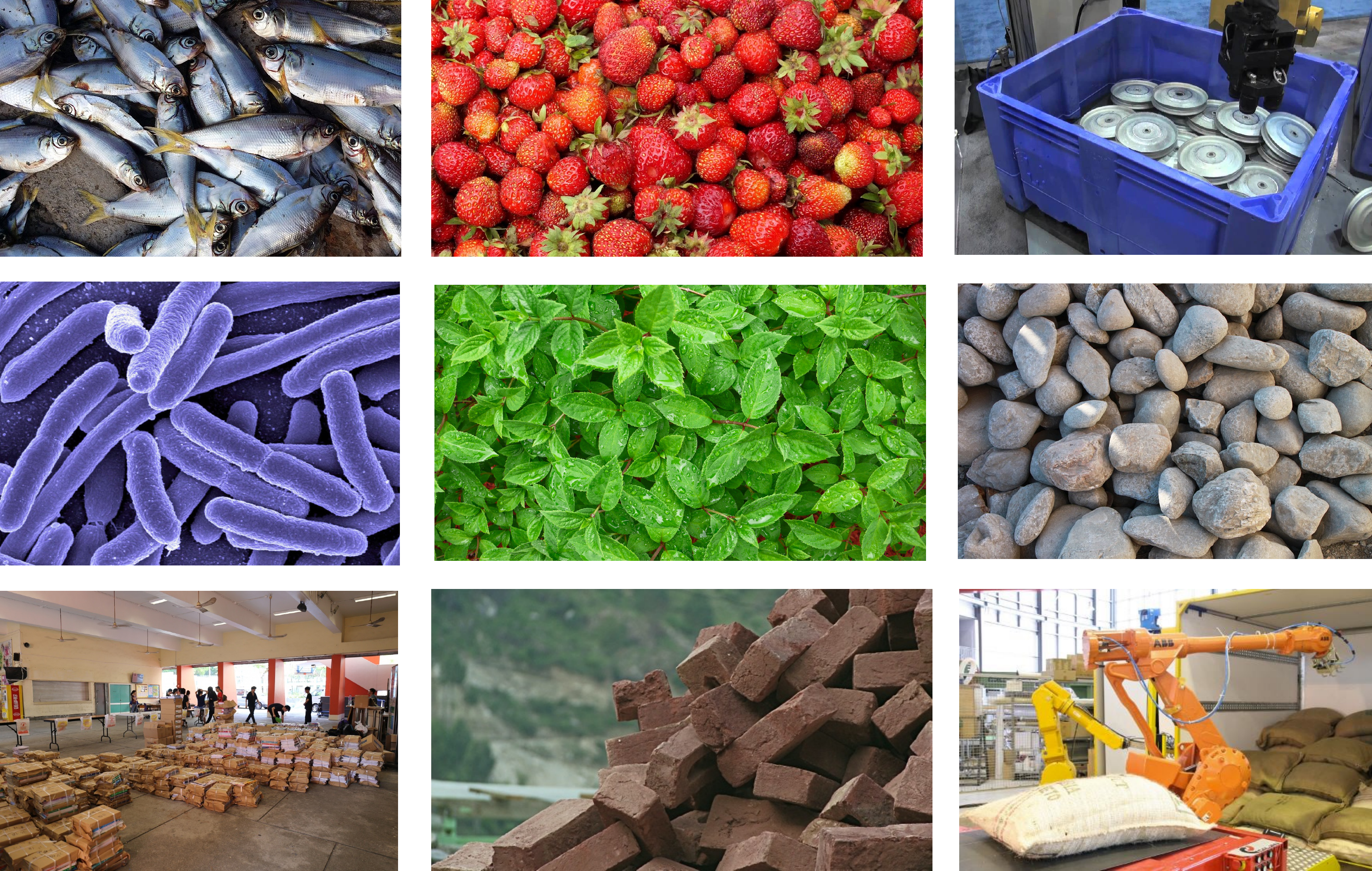}
  \caption{Typical homogeneous object clusters.}
  \label{fig:scenario case}
  \end{center}
\end{figure}

Homogeneous object clusters (HOC) are ubiquitous. From microscopic cells to gigantic galaxies, they tend to cluster together. Figure~\ref{fig:scenario case} shows typical examples. Delineating individual homogeneous objects from their cluster gives an accurate estimate of the number of instances which further enables many important applications: for example, in medicine, various blood cell counts give crucial information on a patient's health. 

Instance segmentation for HOCs is by no means a trivial task despite the uniformity of the target objects.  Directly applying current best performing instance segmentation methods~\cite{he2017mask,li2016fully}, many of which are based on some kind of Deep Convolutional Neural Network (DCNN), meets a bottleneck: unaffordable annotation cost. All of these segmentation models require a large number of annotated images for training purpose. Unlike typical object segmentation (e.g., car, chair), homogeneous clustered objects are densely distributed in a given image with various degrees of occlusion. Pixel-wise labeling these objects is extremely time consuming. However, in many realistic scenarios (e.g., merchandise sold in a supermarket), we have tens of thousands of categories to process. Category-specific annotation is impractical to the instance segmentation problem we address in this paper. We need to automatically generate large amount training data (HOC images with segmentation annotation) with cheap cost.

Generative adversarial nets \cite{goodfellow2014generative} has been widely used to generate images and is a seemingly promising direction. However, the GAN framework cannot generate images with pixel-level annotation, so the segmentation models mentioned above cannot be trained using such images. RenderGAN \cite{sixt2016rendergan} is proposed to generate images from labels. However the method also cannot generate images with  pixel-level annotation. Image-to-image translation \cite{isola2016image} based on conditional GAN \cite{mirza2014conditional} can get the annotation from images but it requires a large collection of image-annotation pairs to train, and thus it still needs a lot of laborious annotation.

Driven by the above considerations, in this paper, we propose a novel framework to tackle the challenging instance segmentation problem.
Inspired by~\cite{oneshot}, our learning framework is \textbf{one-shot} because it learns by looking only once the single sample captured in a single short video, which avoids the cumbersome collection of large-scale image datasets for training. Then, these single-object video frames are used to automatically synthesize realistic images of homogeneous object clusters. In doing so, we can acquire a large amount of training data automatically. Therefore, our framework is \textbf{annotation-free}. However, generating visually realistic images is a non-trivial task, since structural constraint (i.e., cluster layout should look reasonable) and illumination should be taken into consideration. In this paper, we propose a novel image synthesis framework to capture key priors from real images depicting HOCs. Structure prior is captured by learning a structured likelihood function. We generate structurally realistic synthetic images of HOC by placing synthetic objects in a specific way that maximizes the structured likelihood given by our defined function. Illumination is simulated through an efficient illumination transformation method we develop to transform both synthetic images and real images to share a similar illumination condition.

\begin{figure}
  \begin{center}
    \includegraphics[width=\linewidth]{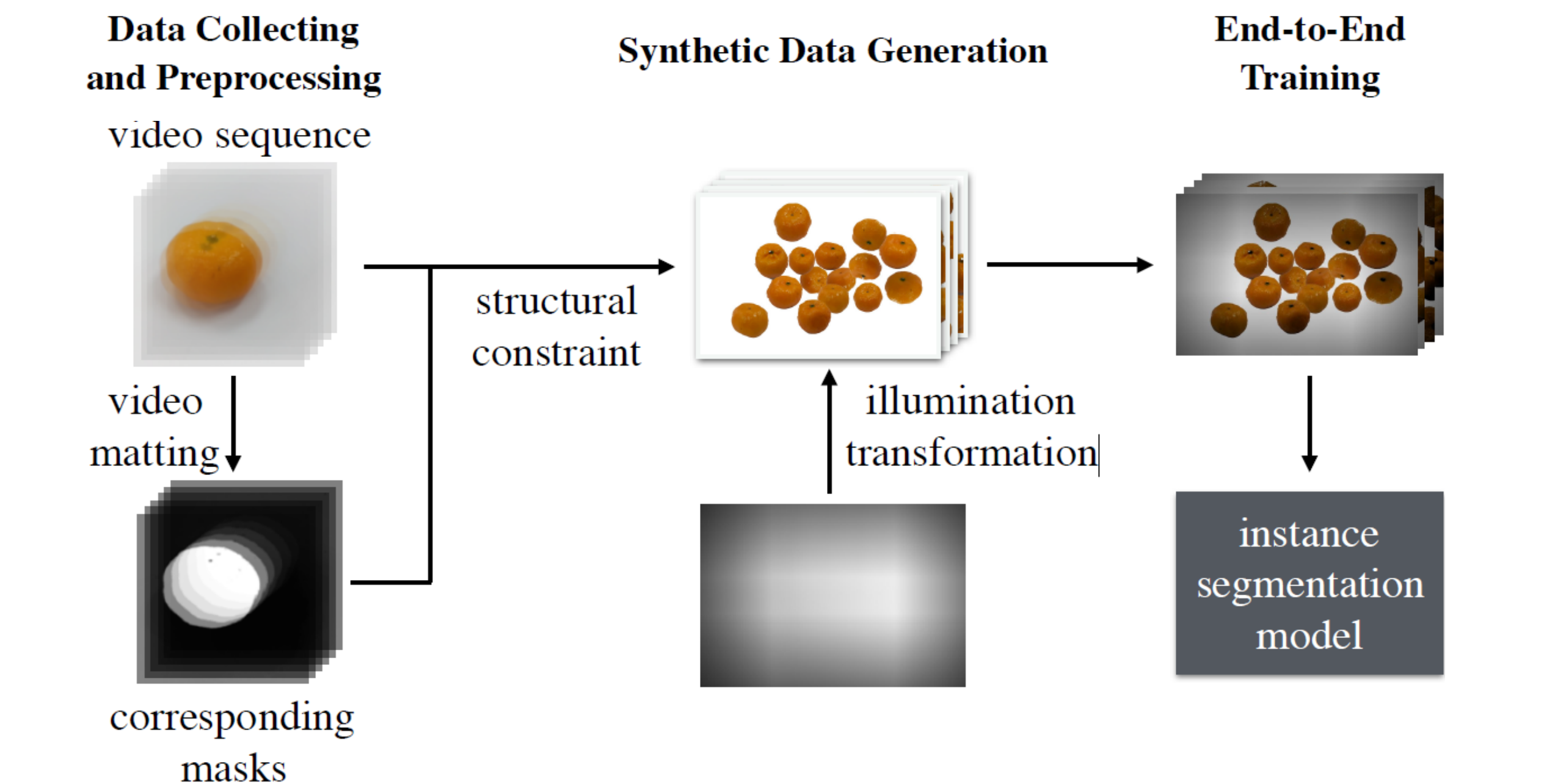}
  \caption{Overview of our pipeline. Our system (1) takes single-object video as input, extracts the mask for each frame, (2) generates synthetic images of homogeneous objects in cluster, then (3) uses the synthetic images to train an instance segmentation model.}
    \label{fig:framework}
  \end{center}
\end{figure}

To benchmark our performance, a dataset is built that consists of 200 properly selected and annotated images of homogeneous clustered objects. The extensive experiments show that our approach significantly improves the segmentation performance over baseline methods on the proposed dataset.

The contributions of this paper are summarized as follows:
\begin{enumerate}
\item We propose an efficient framework for instance segmentation of HOCs.
Our proposed method significantly reduces the cost of data collection
and labeling.
\item We propose an efficient method to generate realistic synthetic training data which significantly improves instance segmentation performance.
\item We build a dataset consisting of HOC images.  The dataset is used to evaluate our proposed method. The dataset and codes will be published with the paper.
\end{enumerate}

\section{Related Work}

Instance segmentation, which aims to assign class-aware and instance-aware label to each pixel of the image, has witnessed rapid progress recently in computer vision. A series of work reporting good performance has been proposed~\cite{he2017mask,li2016fully}, in which all of them use some kind of DCNN trained on large datasets~\cite{mscoco,cityscapes}. However, none of these current representative instance segmentation benchmarks has adequately addressed the task of segmenting HOCs at the instance level. Unlike the classical instance segmentation problem, the testing images for our problem have much more homogeneous instances on average, and exhibit a much higher degree of occlusion. Since currently all of the best models in performing instance segmentation require a large volume of pixel-level annotated images during training, and that no public dataset is available for our specific task, the traditional training framework is inadequate to address our problem of HOC instance segmentation.

Our work is inspired by one-shot learning~\cite{oneshot}. One-shot learning learns object categories from one training sample. Recently, it has been applied to solve image segmentation problem~\cite{CaellesMPLCG16,rong2016one}. \cite{CaellesMPLCG16} use intrinsic continuity to segment objects in a video, but pixel-level annotation for the first frame is required. The work of~\cite{rong2016one} addresses the problem of segmenting gestures in video frames by learning a probability distribution vector (PDV) that describes object motions. However, these methods still require a number of images with pixel-level annotation.

In this paper we use single-object images to synthesize images of HOC as our training data. Recently, many studies have been done to use different methods to generate synthetic images to train a DCNN model. In~\cite{papon2015semantic} realistic synthetic scenes are generated by placing object models at random in a virtual room, and uses the synthetic data to train a DCNN. In the area of object detection, the authors in \cite{peng2015learning} propose to use 3D CAD models to generate synthetic training images automatically to train an object detector, while in 3D recognition,  the work in \cite{su2015render} synthesizes images by overlaying images rendered from 3D model on top of real images.  Among these methods using synthetic data generation, \cite{papon2015semantic} is most similar to ours (i.e., synthesizing images by placing objects on a background sequentially). While they simply place the synthesized objects randomly, we propose to learn how to place synthesized objects based on the knowledge learned from realistic images of HOCs.

\section{Our Method}
\subsection{Data Collection and Preprocessing}
\label{Our Method:Data Collecting and Preprocessing}
We aim to collect single-object images quickly and efficiently, so videos are the natural choice. We capture a short video for a single object, which is processed to extract individual frames alongside with the corresponding masks. The details are described in the following.

\subsubsection{One-Shot Video Collection}
Suppose we want to segment each orange from a cluster of oranges. A common solution is to collect a large number of images of orange clusters in different layouts, followed by annotating all of them and then training an instance segmentation model using the annotated images. In this paper, instead of adopting such conventional data collection, we perform the following: put one single orange at the center of a contrastive background, and take a video of the orange at different angles and positions. Such video typically lasts for about 20 seconds. It takes only a few minutes to acquire the data we require, which significantly reduces the cost of data collection compared to the previous methods. Since in our framework this single short video capturing one single object is all we need for learning, our framework may be regarded as a close kin to \textbf{one-shot learning} (i.e., learning by looking only once, one video in our case).

\subsubsection{Video Matting}
\begin{figure}[h]
\vspace{1mm}
  \begin{center}
  	\includegraphics[width=\linewidth]{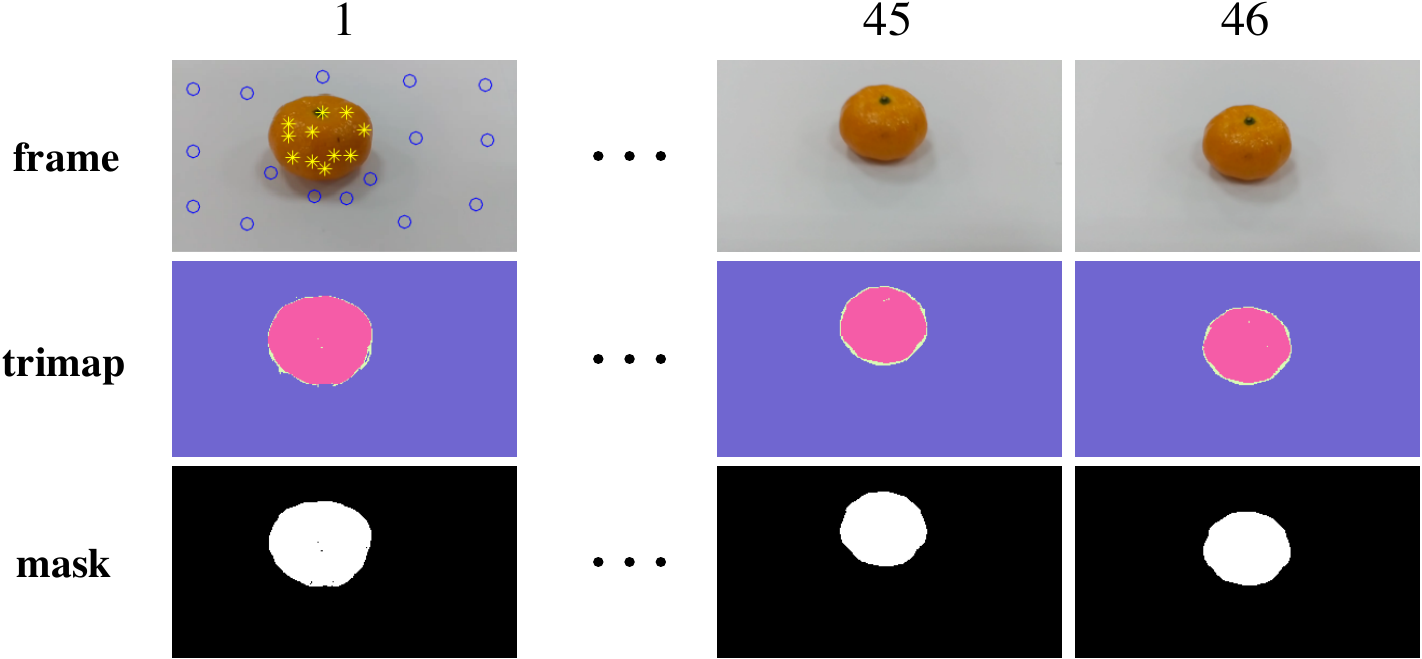}
  	\caption{Example of our video processing method. For the first frame, seeds of foreground and background are automatically sampled based on color and location priors. Trimaps are interpolated across the video volume using optical flow. Matting technique uses the flowed trimaps to yield high-quality masks of the moving orange.}
  	\label{fig:data preprocess}
  \end{center}
\end{figure}

Our goal here is to automatically obtain the mask of each single-object image (i.e., the video frames) in an annotation-free manner. Since the foreground and background are controlled, and the image object is at the center, we apply color and location prior in the first frame to automatically sample seeds of the object and background respectively. Then we take the seeds as input and apply KNN matting~\cite{Chen:2012:KM} on the video sequence to produce the mask of each frame.  We take into consideration temporal coherence, and instead of producing a hand-drawn trimap (a pre-segmented image consisting of three regions: foreground, background and unknown) for every single frame, classical optical flow algorithm is applied for trimap generation and propagation. Figure~\ref{fig:data preprocess} shows an example.

\begin{figure*}[t]
\vspace{1mm}
  \begin{center}
  	\includegraphics[width=\linewidth]{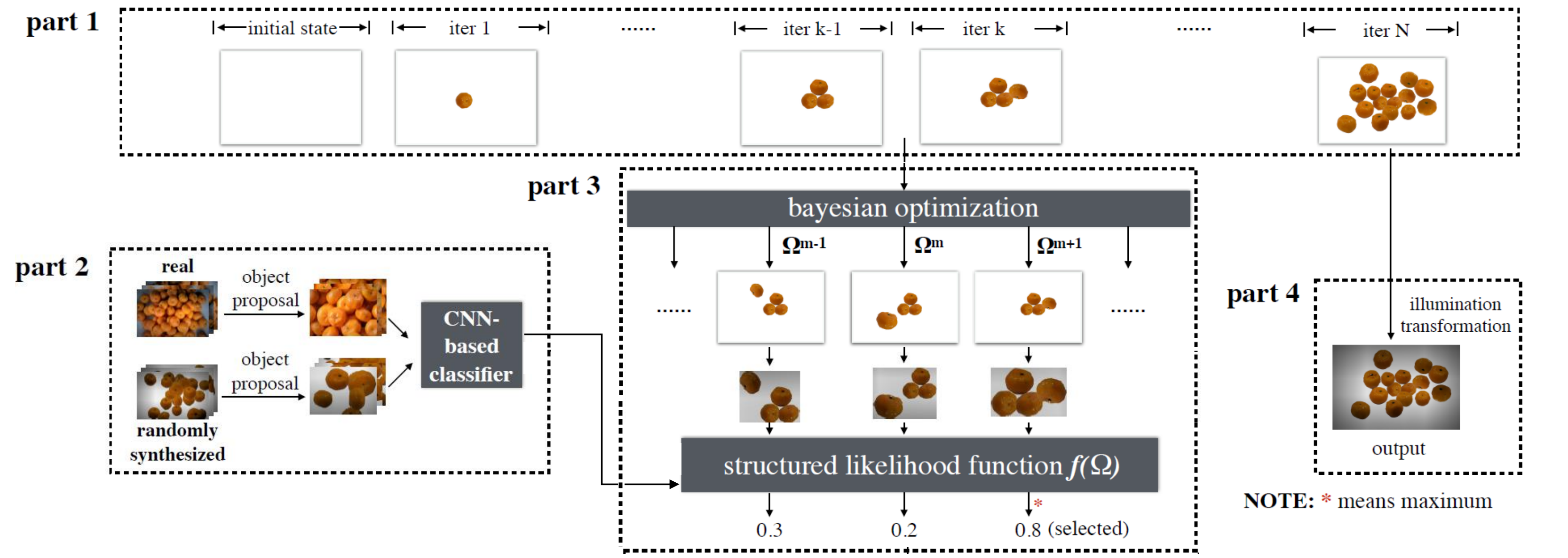}
  	\caption{Synthetic images generation. $N$ objects are selected and placed on a white background sequentially in an iterative manner (\textbf{part 1}). We start by placing one object directly on the background in iteration 1. For iteration $k, 1<k \le N$, we use bayesian optimization (\textbf{part 3}) to approximate an optimal placement that is guided by our structured likelihood function (\textbf{part 2}). During bayesian optimization process, $M$ points $\{\Omega^1,...,\Omega^M\}$ are sampled, while we only plot three due to space constraint. Each point is scored by $f(\Omega)$. The optimal placement is the point with the highest score. Then we apply illumination transformation (\textbf{part 4}) on the output of iteration $N$ to obtain the final result.}
  	\label{fig:synthetic overview}
  \end{center}
\end{figure*}

\subsection{Synthetic Data Generation}
\label{Our Method:Synthetic data generation}
Given the segmented single-object images from the previous stage, we generate synthetic images in realistic structure and illumination. Realistic images of HOC are generated in an iterative manner, followed by illumination transformation to make synthetic images and real images share a similar illumination condition.

\subsubsection{Structural Constraint}
\label{Our Method:Synthetic data generation:Structural constraints}

\paragraph{Structured Likelihood Function}
The problem of generating realistic images of HOC deals with finding a solution to properly place $N$ objects
$\{{O}_{1},{O}_{2},...,{O}_{N}\}$
sequentially in order to maximize its likelihood of being a real image of HOC. Let $\textbf{I}_k$ denote the image we obtain after placing $k$ objects $\{{O}_{1},...,{O}_{k}\}, 1 < k \le N$, and $P(\cdot)$ denote the likelihood of being a real image. Our problem is equivalent to finding a solution to maximize $P(\textbf{I}_N)$.

The image we get after placing $k$ objects, $\textbf{I}_k$, is determined by three factors: the images we get after placing $k-1$ objects $\textbf{I}_{k-1}$, the $k$-th chosen object $O_k$ and the way we place $O_k$ in $\textbf{I}_{k-1}$. We model our object placement by a four-parameter operation set $\Omega = \{\theta, \gamma, x, y\}$, in which $\theta$ denotes the rotation, $\gamma$ denotes the resize factor, and $x$ and $y$ denote the coordinates of the center of $O_k$ in $\textbf{I}_{k-1}$. The relation between $\textbf{I}_k$ and $\textbf{I}_{k-1}$, $O_k$, $\Omega_k$ can be written as:
\begin{equation}
	\textbf{I}_k=g(\textbf{I}_{k-1},O_k,\Omega_k)
\end{equation}


Given a sequence of $\{{O}_{1},{O}_{2},...,{O}_{N}\}$, our goal is to find a $(\Omega_1,...,\Omega_N)$ that maximizes $P(\textbf{I}_N)$. Since $\{{O}_{1},{O}_{2},...,{O}_{N}\}$ is given, finding a solution to maximize $P(\textbf{I}_N)$ is equivalent to finding an optimal $(\Omega_1,...,\Omega_N)$ that maximizes $P(\textbf{I}_N)$. Since searching for an optimal $(\Omega_1,...,\Omega_N)$ is intractable, we simplify the problem by applying greedy search algorithm, that is, in each iteration, the operation that maximizes $P(\textbf{I}_k)$ is applied. Given $\textbf{I}_{k-1}$ and $O_k$ $(1<k \le N)$, we search for an optimal $\Omega_k$ that maximizes $P(\textbf{I}_k)$:
\begin{equation}
\begin{aligned}
	\overline{\Omega}_k &=\arg\max_{{\Omega}_{k}\in \mathcal{D}} P(\textbf{I}_k)\\
	&=\arg\max_{ {\Omega}_{ k }\in \mathcal{D} } P(g(\textbf{I}_{k-1},O_k,\Omega_k))
\end{aligned}
\end{equation}
where $\mathcal{D}$ is the feasible set of object placement.  When $\textbf{I}_{k-1}$ and $O_k$ are given, $P(g(\textbf{I}_{k-1},O_k,\Omega_k))$ is the function of $\Omega_k$. Let $f(\Omega_k)=P(g(\textbf{I}_{k-1},O_k,\Omega_k))$, then
\begin{equation}
	\overline{\Omega}_k=\arg\max_{ {\Omega}_{ k }\in  \mathcal{D} } f(\Omega_k;\textbf{I}_{k-1},O_k)
\end{equation}
We aim to find an optimal solution $\overline{\Omega}_k$ that maximizes $f(\Omega_k)$. Since $f(\Omega_k;\textbf{I}_{k-1},O_k)=P(I_k)$, $f(\Omega_k;\textbf{I}_{k-1},O_k)$ represents the likelihood of $\textbf{I}_k$ being a real image given $\textbf{I}_{k-1}$ and $O_k$. We turn to DCNN to judge whether $\textbf{I}_{k-1}$ is real or otherwise. In detail, we build the classifier by finetuning a pre-trained ImageNet model, where we use object proposals of real images as positive samples, and object proposals of randomly synthesized images (i.e., images synthesized by placing objects at random) as negative samples. We take the tight bounding box~\cite{su2012crowdsourcing} of $\textbf{I}_{k}$ as the input of the learnt classifier. The softmax output of the real class is taken as the output of $f(\Omega_k;\textbf{I}_{k-1},O_k)$.

\paragraph{Bayesian optimization framework}
Since $f(\Omega_k)$ is a black-box function (i.e., we do not have a specific expression of the function), we cannot optimize $f$ by computing its derivative. We solve this problem by adopting bayesian optimization~\cite{bayesian}. Bayesian optimization is a powerful strategy for optimization of black-box function.

Here, we use bayesian optimization to optimize our continuous objective function $f(\Omega_k)$. Bayesian optimization behaves in an iterative manner. At iteration $m$, given the observed point set $D_{m-1}=\{(\Omega^1_k,f(\Omega^1_k)),...,(\Omega^{m-1}_k,f(\Omega^{m-1}_k))\}$, we model a posterior function distribution by the observed points. Then, the acquisition function (i.e., a utility function constructed from the model posterior) is maximized to determine where to sample the next point $(\Omega^{m}_k,f(\Omega^{m}_k))$. $(\Omega^{m}_k,f(\Omega^{m}_k))$ is collected and the process is repeated. Iteration ends when $m=M$ ($M$ is a user-defined parameter). For more detail on optimization framework please refer to \cite{bayesian}.

\paragraph{Overall algorithm}
We generate our realistic HOC images by placing objects sequentially in an iterative manner (see Figure \ref{fig:synthetic overview}). We start by directly placing one object on a white background in iteration 1. In iteration $k$ ($1<k\le N$), given $\textbf{I}_{k-1}$ and $O_k$, bayesian optimization is applied to approximate an optimal $(\theta_k,\gamma_k,x_k,y_k)$ that maximizes the structured likelihood function $f(\Omega_k;\textbf{I}_{k-1},O_k)$. Iteration ends when $k=N$. $N$ is a prior given by user. User provides a range of how many instances per image in his (or her) case, thus, $N$ is a random integral in this range. For example, in our dataset, $N$ is a random integral between 10 and 30.

\subsubsection{Illumination Transformation}
\label{Illumination transformation}
To simulate lighting condition, we develop an efficient method to transform the synthetic and the real image so that they share a similar illumination condition.

We are inspired by~\cite{leong2003correction} where an uneven illumination correction method was proposed using Gaussian smoothing. We first convert both the synthetic image and the real image from RGB color space to HSV space, where $V$ represents illumination information. Then, we implement detail removing, by using a large kernel Gaussian smoothing on both images to model general illumination condition. This general illumination condition (denoted as $blur(V_{real})$) is imposed on both real and synthetic images (after mean subtraction) to unify illumination.
\begin{equation}
	V_{syn}=V_{syn}-\mbox{mean}(V_{syn})+blur(V_{real})
\end{equation}
\begin{equation}
	V_{real}=V_{real}-\mbox{mean}(V_{real})+blur(V_{real})
\label{eq:illumination}
\end{equation}
Here, $V_{syn}$ and $V_{real}$ respectively denote the V channel of the synthetic image and the real image, and $\mbox{mean}(\cdot)$ denotes the global mean value of 2-D matrix.

\begin{figure}[!htp]
  \begin{center}
  	\includegraphics[width=\linewidth]{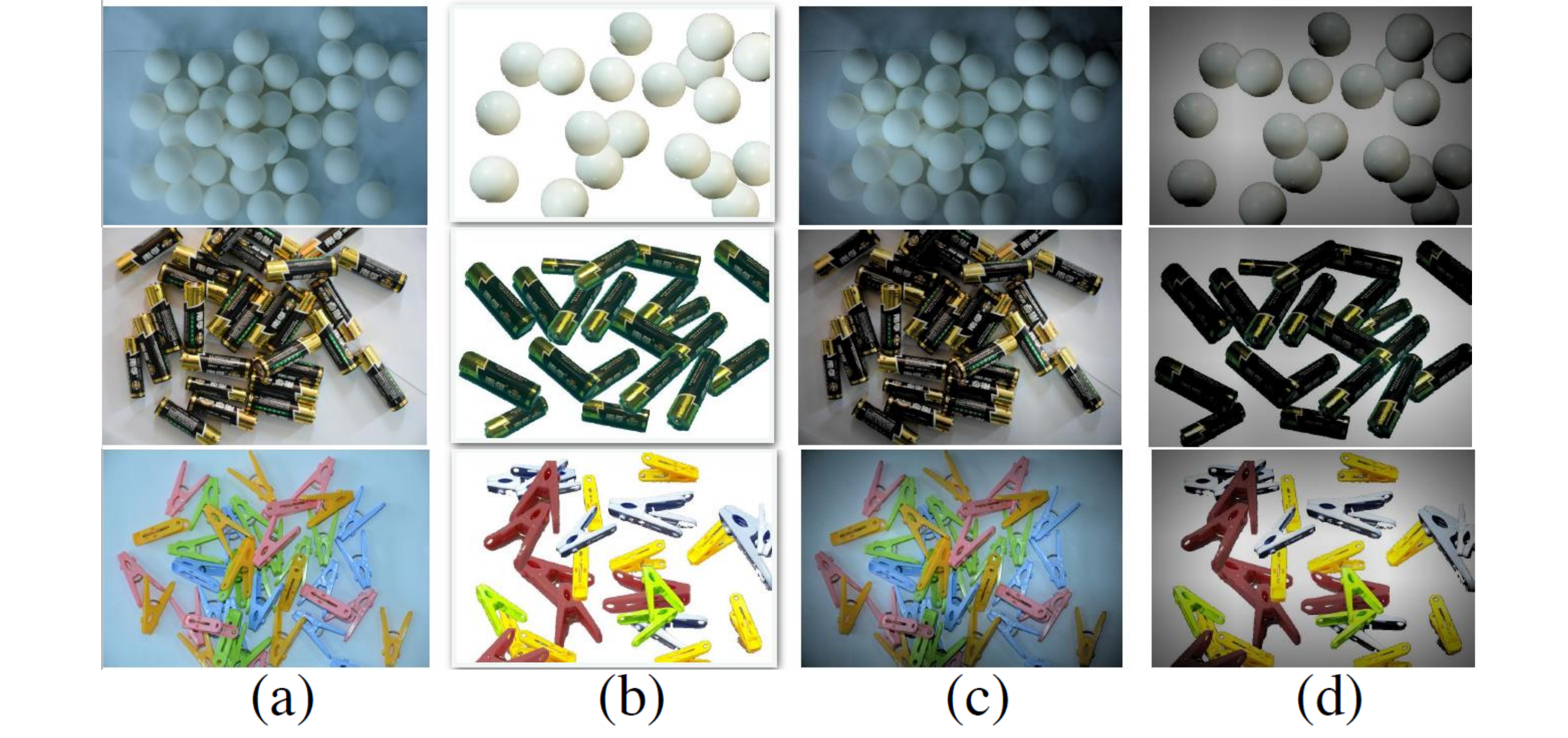}
  \caption{Examples of our illumination transformation method. (a) real image before illumination transformation, (b) synthetic image before illumination transformation, (c) real image after illumination transformation, (d) synthetic image after illumination transformation.}
  \label{fig:illumination_result}
  \end{center}
\end{figure}

Finally, we convert the synthetic image and real image from HSV space back to RGB space. Figure~\ref{fig:illumination_result} shows that our illumination transformation algorithm makes real and synthetic images share the similar illumination condition.

\vspace{-5mm} 
\subsection{End-to-End Training}
Finally, we use the synthetic images together with their corresponding annotations generated by the above method as our training data, and train an instance segmentation model in an end-to-end manner.

In this paper, we adapt Mask R-CNN~\cite{he2017mask} to handle our task. We train the model using our synthetic data. In testing phase, the model works on real image. Before the testing image is fed to the model, we adjust its illumination using Eq. \ref{eq:illumination} to ensure its illumination condition is similar to training images.

\begin{table*}[htp]
\centering
\caption{Results on mAP$^r$@0.5 on our dataset. All numbers are percentages \%.}
\label{table1}
\tabcolsep=0.11cm
\begin{tabular}{c|cccccccccc|c}
&badminton&battery&clothespin&grape&milk&hexagon nut&orange&ping pong&tissue&wing nut&mAP \\ \hline
\textbf{Single} &12.1&23.2&4.4&19.3&8.2&17.5&17.6&14.2&13.1&21.1&15.1 \\
\textbf{Random} &40.6&50.9&38.4&50.8&26.7&52.9&63.8&67.2&83.9&32.9&50.8 \\
\textbf{Random+illumination} &44.6&48.7&34.3&41.6&26.0&46.3&54.9&64.2&68.9&39.4&46.9 \\
\textbf{Random+structure} &34.2&39.3&52.6&\textbf{72.7}&31.3&62.8&\textbf{90.3}&\textbf{81.7}&\textbf{90.7}&23.7&57.9 \\

\textbf{Ours} &\textbf{53.0}&\textbf{69.5}&\textbf{67.7}&72.5&\textbf{52.6}&\textbf{73.6}&90.0&81.2&90.4&\textbf{48.4}&\textbf{69.9}
\end{tabular}
\centering
\caption{Results on mAP$^r$@[0.5:0.95] on our dataset. All numbers are percentages \%.}
\label{table2}
\tabcolsep=0.11cm
\begin{tabular}{c|cccccccccc|c}
&badminton&battery&clothespin&grape&milk&hexagon nut&orange&ping pong&tissue&wing nut&mAP \\
\hline
\textbf{Single} &8.7&21.2&2.0&16.8&5.0&15.2&16.0&11.8&9.9&18.8&12.5 \\
\textbf{Random} &33.1&44.7&29.3&44.8&20.9&43.4&56.3&59.5&68.4&27.7&42.8 \\
\textbf{Random+illumination} &36.5&43.2&28.4&37.2&20.8&38.3&50.4&56.4&56.8&35.5&40.4 \\
\textbf{Random+structure} &29.8&36.5&36.7&\textbf{66.5}
&22.0&45.2&83.8&\textbf{76.4}&\textbf{80.4}&20.7&49.8 \\
\textbf{Ours}  &\textbf{44.2}&\textbf{60.3}&\textbf{46.2}&65.4&\textbf{41.3}&\textbf{54.8}&\textbf{84.0}&75.6&\textbf{80.4}&\textbf{39.3}&\textbf{59.2} \\

\end{tabular}
\end{table*}

\section{Experimental Evaluation}
\subsection{Dataset}
Our task is totally new and none of current datasets is suitable for evaluating this problem. Our dataset is designed for benchmarking methods for instance segmentation of HOCs. Our dataset contains 10 types of objects; some are solid (\textit{e.g.}, ping pong ball) while some are fluid (\textit{e.g.}, bagged milk). For each type of object, 20 images are taken under different illumination conditions. We annotate all of the instances with less than 80\% occlusion. Our dataset has $3,669$ instances in total, each image has 18.3 instances on average. We do not build an extremely larger dataset due to budget constraints, but it is sufficient to benchmark this problem.

\subsection{Implementation details}

\paragraph{Structured likelihood function} In the training step, we construct the likelihood function by finetuning the AlexNet~\cite{alexnet} pre\-trained model. For each type of object, we use \cite{edgeBoxes} to generate object proposals for both real images and randomly synthesized images (i.e., images synthesized by placing objects at random), and filter out proposals whose area are less than 0.01 of the original image. Then, we randomly choose 30,000 proposals from real images as positive examples, and 30,000 proposals from synthetic images as negative examples.  We finetune the model for 30,000 iterations with a learning rate of 0.001, minibatch size of 32, momentum of 0.9 and weight decay of 0.01.

In the inference step, given $\textbf{I}_{k-1}$, $O_k$ and $(\theta_k,\gamma_k,x_k,y_k)$, we rotate $O_k$ by $\theta_k$ degrees, resize it by a factor of $\gamma_k$, and place the center of $O_k$ after rotation and resizing at location $(x_k,y_k)$. When we place $O_k$, it can only be overlapped by previous objects. Then the tight bounding box of the new image we get after placing $O_k$ is taken as input of the finetuned model. The softmax output of the real class is taken as the output of our structured likelihood function.

\paragraph{Instance segmentation model} We adopt the Mask R-CNN~\cite{he2017mask} as our instance segmentation model. We finetune the VGG-16 model~\cite{vgg16} using the synthetic data generated by our method for 20,000 iterations with a learning rate of 0.002, minibatch size of 4, momentum of 0.9 and weight decay of 0.0001 on four Titan X GPU. Other settings are identical with~\cite{he2017mask}.

\subsection{Results and analysis}
We follow the protocols used in \cite{Hariharan2014Simultaneous,he2017mask,li2016fully} for evaluating our proposed method for solving the problem of instance segmentation of HOCs. We evaluate using standard COCO evaluation metric, mAP$^r$@[0.5:0.95], as well as traditional mAP$^r$@0.5 metric. Table \ref{table1} and Table \ref{table2} respectively show the results on the mAP$^r$@0.5 and the mAP$^r$@[0.5:0.95] metrics on our dataset.  

\begin{enumerate}
\item \textbf{Single} is our baseline method, which uses outputs from Section \ref{Our Method:Data Collecting and Preprocessing} (i.e., images containing one single object and corresponding annotation) as training data to train the Mask R-CNN~\cite{he2017mask}. The baseline method reports mAP$^r$@0.5 of $15.1\%$ and mAP$^r$@[0.5:0.95] of $12.5\%$. Since the training images of \textbf{Single} only contain a single object each, while testing realistic images contain multiple instances, the poor result is within our expectation.

\item \textbf{Random} is the result of training with randomly synthesized images (i.e., images synthesized by placing objects at random), which achieves mAP$^r$@0.5 of $50.8\%$ and mAP$^r$@[0.5:0.95] of $42.8\%$. This is a large improvement over \textbf{Single}, although \textbf{Single} is not designed for the task, while the comparison is indicative it is somewhat unfair.

\item \textbf{Random+illumination} applies the illumination transformation method in Section \ref{Illumination transformation} to transform synthetic images generated in \textbf{Random} and real images to share similar illumination condition. It uses synthetic images after illumination transformation as training data to train Mask R-CNN~\cite{he2017mask} model. The result shows a slight decrease in performance using the mAP$^r$@0.5 metric (of about $3.9\%$) and mAP$^r$@[0.5:0.95] metric (of about $2.4\%$) over \textbf{Random}, indicating that illumination transformation does not help when training with randomly synthesized images.

\item \textbf{Random+structure} uses the method proposed in Section~\ref{Our Method:Synthetic data generation:Structural constraints} to generate structurally realistic synthetic images {\em without} illumination transformation as training data. The result shows a $7.1\%$ improvement on mAP$^r$@0.5 and a $7.0\%$ improvement on mAP$^r$@[0.5:0.95] over \textbf{Random}. The significant improvement demonstrates the efficacy of our method which synthesizes realistic HOC images.

\item \textbf{Ours} is our method which trains the model using synthetic images generated under both the structural constraint and with illumination transformation. Our method reports $12.0\%$ higher in mAP$^r$@0.5 and $9.4\%$ higher in mAP$^r$@[0.5:0.95] over \textbf{Random+structure}. While illumination transformation does not show effectiveness for training with randomly synthesized images, it indeed significantly improves performance for training with structurally realistic synthetic images.
\end{enumerate}
\begin{figure}[!htb]
  \begin{center}
  	\includegraphics[width=\linewidth]{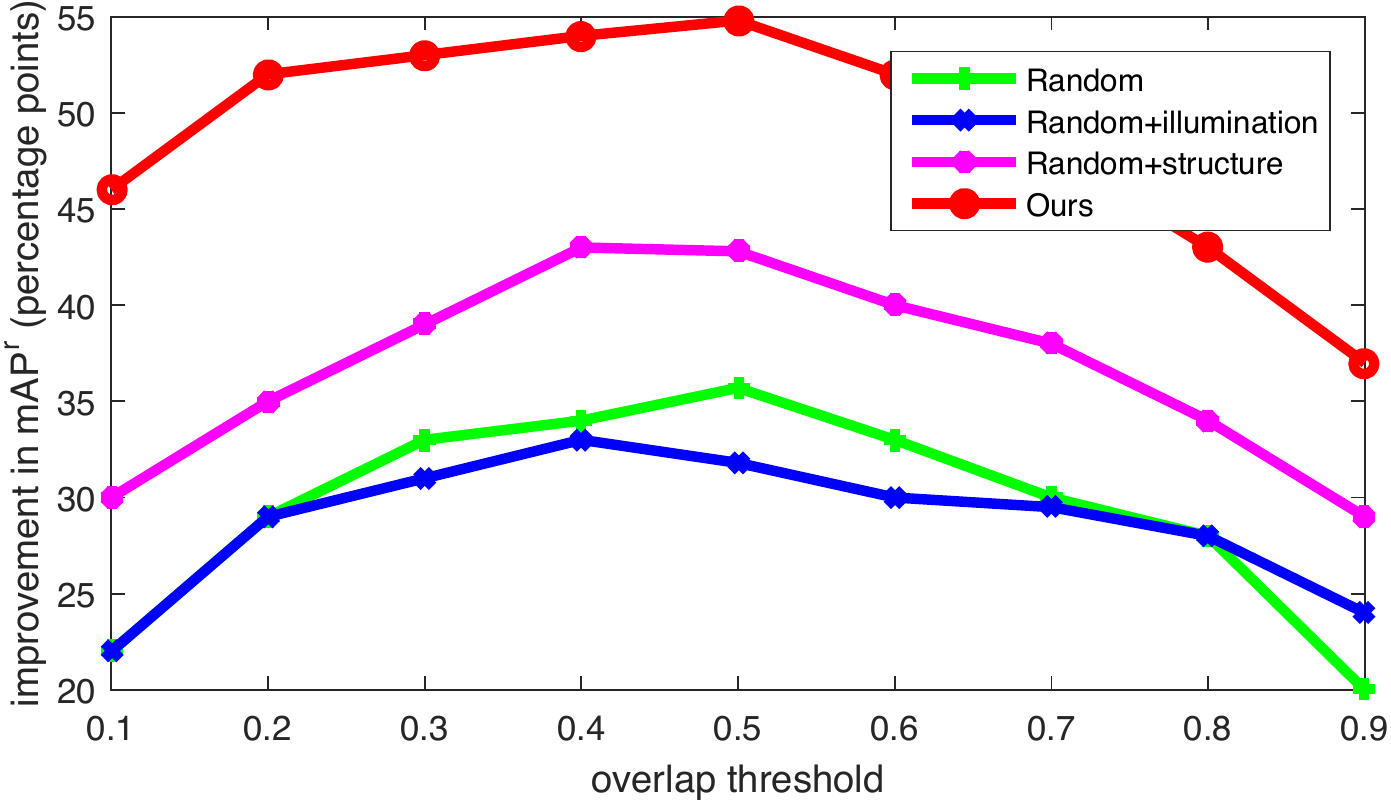}
  	\caption{Improvement in mAP$^r$ over \textbf{Single} for a variety of overlap thresholds. Our method reports improvements on all thresholds.}
  	\label{fig:experiment result}
  \end{center}
\end{figure}
\vspace{-3mm}
 \begin{figure}[htp]
  \begin{center}
  	\includegraphics[width=0.9\linewidth]{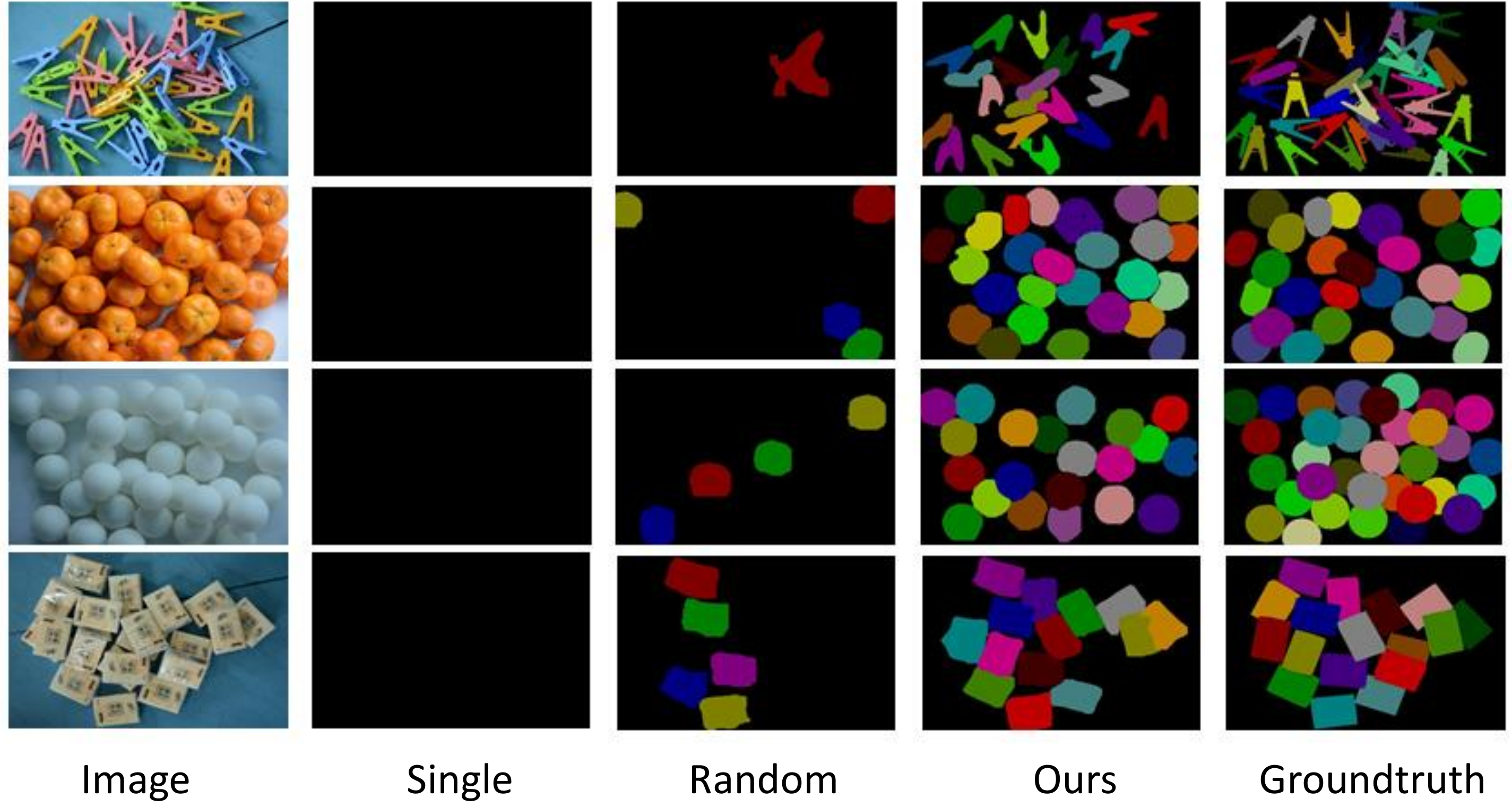}
  \end{center}
   \caption{Qualitative segmentation results on our dataset. Each color denotes one instance. Our method can segment more instances than other baseline methods.}
  \label{fig:qualitative result}
\end{figure}

Following the evaluation in~\cite{Hariharan2014Simultaneous}, Figure~\ref{fig:experiment result} plots the improvements on mAP$^r$ over \textbf{Single} at 9 different IoU thresholds. As shown in Figure \ref{fig:experiment result}, the performance of our proposed method is superior to other methods at every threshold, which demonstrates the robustness of our algorithm. We also show some qualitative results in Figure \ref{fig:qualitative result}, where we can observe that \textbf{Random} (trained with images synthesized by placing objects at random) can only segment objects that are not occluded by other objects, while our method can segment out a much larger number of partially occluded instances. This demonstrates that our image synthesis method is able to learn the occlusion pattern of real images. Besides, as shown in Table \ref{table1} and Table \ref{table2}, the significant gap between \textbf{Random+structure} and \textbf{Ours} ($12.0\%$ in mAP$^r$@0.5 and $9.4\%$ in mAP$^r$@[0.5:0.95]) indicates the effectiveness of our illumination transformation algorithm.

\subsection{Comparison of training with existing dataset}
\begin{figure}[!htb]
  \begin{center}
  	\includegraphics[width=\linewidth]{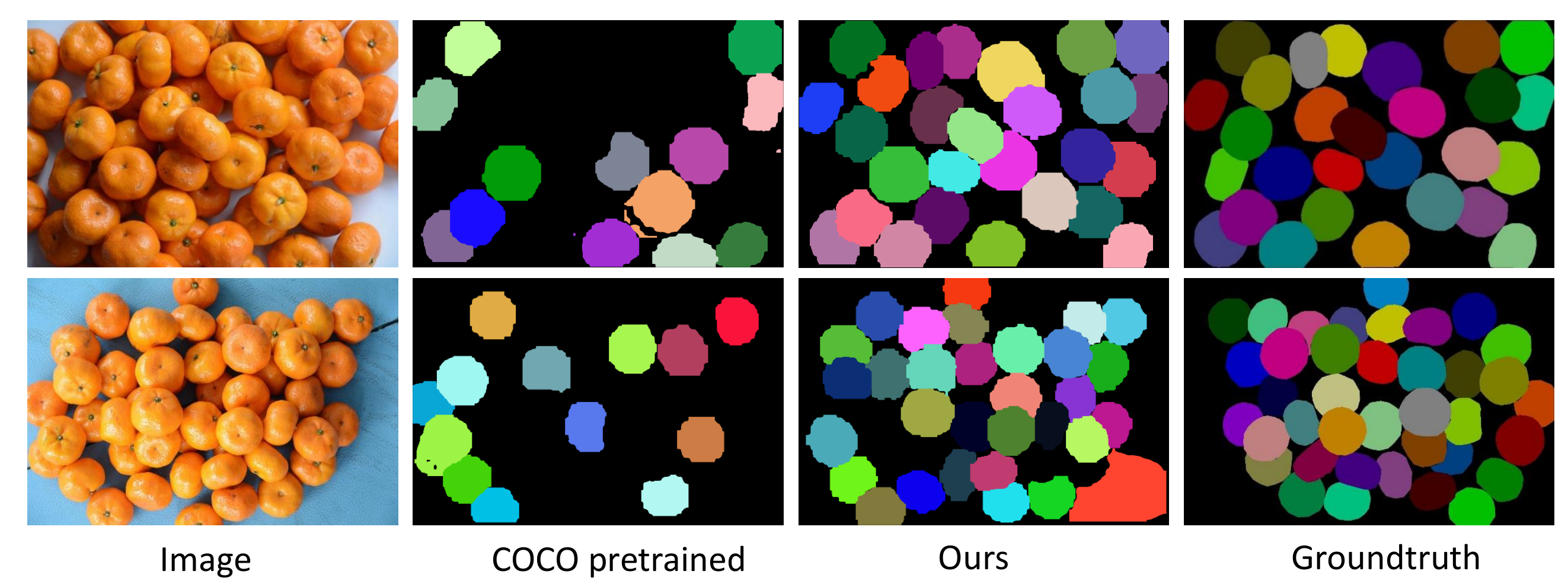}
  	\caption{Comparison of COCO pre-trained model and our method. Our method is better at handling occlusion.}
  	\label{fig:qualitative result 2}
  \end{center}
\end{figure}

To further demonstrate the efficacy of our method which synthesizes HOC images is to compare our model with one that is trained on equal number of real HOC images with instance segmentation annotation. However, due to our limited budget, we cannot afford to build such a large dataset of HOC images with thorough annotation.  Alternatively, we validate the effectiveness of our proposed method by comparing with state-of-the-art an instance segmentation model trained on existing dataset. Since our proposed dataset of HOCs and the COCO dataset have \textit{orange} class in common, we use Mask R-CNN trained on COCO to directly test for object categories in common (i.e., \textit{orange}) on our dataset. Mask R-CNN trained on COCO achieves $86.4\%$ mAP$^r$@0.5 and $72.3\%$ mAP$^r$@[0.5:0.95] on \textit{orange} class in our dataset while our method achieves $90.0\%$ mAP$^r$@0.5 and $84.0\%$ mAP$^r$@[0.5:0.95], which significantly outperforms the baseline method. The reason could be the baseline model does not train on large-scale HOC images, which are difficult to obtain due to expensive human annotation. By contrast, our proposed method can generate high-quality annotated HOC images with little effort which helps to achieve better performance.  
Figure~\ref{fig:qualitative result 2} shows some qualitative results. 

\vspace{-5mm} 
\section{Conclusion and Future Work}
Our work is the first attempt in tapping into learning how to segment homogeneous objects in clusters via single-object samples. The key to our proposed framework is an image synthesis technique based on the intuition that images of homogeneous clustered objects can be synthesized based on priors from real images. We build a dataset consisting of 200 carefully selected and annotated images of homogeneous object of clusters to evaluate our algorithm. While it indeed shows promises, the problem is far from solved. Our current framework may show poor performance on deformable objects (\textit{e.g.}, human) since the collected single-object samples may not cover adequately all the possible appearances of the object, and training with synthetic images generated by these samples makes DCNN prone to overfitting. Nonetheless, given its odds in its present form, we believe this is an interesting work that will spawn future work and useful applications. We plan to produce a larger dataset in the future.

\bibliographystyle{named}
\bibliography{ijcai18}

\begin{thebibliography}{}

\bibitem[\protect\citeauthoryear{Caelles \bgroup \em et al.\egroup
  }{2016}]{CaellesMPLCG16}
Sergi Caelles, Kevis{-}Kokitsi Maninis, Jordi Pont{-}Tuset, Laura
  Leal{-}Taix{\'{e}}, Daniel Cremers, and Luc~Van Gool.
\newblock One-shot video object segmentation.
\newblock {\em CoRR}, abs/1611.05198, 2016.

\bibitem[\protect\citeauthoryear{Chen \bgroup \em et al.\egroup
  }{2013}]{Chen:2012:KM}
Qifeng Chen, Dingzeyu Li, and Chi-Keung Tang.
\newblock Knn matting.
\newblock {\em Pattern Analysis and Machine Intelligence, IEEE Transactions
  on}, 35(9):2175--2188, Sept 2013.

\bibitem[\protect\citeauthoryear{Dai \bgroup \em et al.\egroup }{2015}]{mnc}
Jifeng Dai, Kaiming He, and Jian Sun.
\newblock Instance-aware semantic segmentation via multi-task network cascades.
\newblock {\em CoRR}, abs/1512.04412, 2015.

\bibitem[\protect\citeauthoryear{Everingham \bgroup \em et al.\egroup
  }{2015}]{pascalvoc}
M.~Everingham, S.~M.~A. Eslami, L.~Van~Gool, C.~K.~I. Williams, J.~Winn, and
  A.~Zisserman.
\newblock The pascal visual object classes challenge: A retrospective.
\newblock {\em International Journal of Computer Vision}, 111(1):98--136,
  January 2015.

\bibitem[\protect\citeauthoryear{Fei-Fei \bgroup \em et al.\egroup
  }{2006}]{oneshot}
Li~Fei-Fei, Rob Fergus, and Pietro Perona.
\newblock One-shot learning of object categories.
\newblock {\em IEEE Trans. Pattern Anal. Mach. Intell.}, 28:594--611, 2006.

\bibitem[\protect\citeauthoryear{Geiger \bgroup \em et al.\egroup
  }{2013}]{kitti}
Andreas Geiger, Philip Lenz, Christoph Stiller, and Raquel Urtasun.
\newblock Vision meets robotics: The kitti dataset.
\newblock {\em I. J. Robotic Res.}, 32(11):1231--1237, 2013.

\bibitem[\protect\citeauthoryear{Goodfellow \bgroup \em et al.\egroup
  }{2014}]{goodfellow2014generative}
Ian Goodfellow, Jean Pouget-Abadie, Mehdi Mirza, Bing Xu, David Warde-Farley,
  Sherjil Ozair, Aaron Courville, and Yoshua Bengio.
\newblock Generative adversarial nets.
\newblock In {\em Advances in neural information processing systems}, pages
  2672--2680, 2014.

\bibitem[\protect\citeauthoryear{Handa \bgroup \em et al.\egroup
  }{2015}]{handa2015synthcam3d}
Ankur Handa, Viorica Patraucean, Vijay Badrinarayanan, Simon Stent, and Roberto
  Cipolla.
\newblock Synthcam3d: Semantic understanding with synthetic indoor scenes.
\newblock {\em arXiv preprint arXiv:1505.00171}, 2015.

\bibitem[\protect\citeauthoryear{Hariharan \bgroup \em et al.\egroup
  }{2014}]{Hariharan2014Simultaneous}
Bharath Hariharan, Pablo Arbel\'{a}ez, Ross Girshick, and Jitendra Malik.
\newblock Simultaneous detection and segmentation.
\newblock In {\em European Conference on Computer Vision}, pages 297--312,
  2014.

\bibitem[\protect\citeauthoryear{He \bgroup \em et al.\egroup
  }{2017}]{he2017mask}
Kaiming He, Georgia Gkioxari, Piotr Doll{\'a}r, and Ross Girshick.
\newblock Mask r-cnn.
\newblock {\em arXiv preprint arXiv:1703.06870}, 2017.

\bibitem[\protect\citeauthoryear{Isola \bgroup \em et al.\egroup
  }{2016}]{isola2016image}
Phillip Isola, Jun-Yan Zhu, Tinghui Zhou, and Alexei~A Efros.
\newblock Image-to-image translation with conditional adversarial networks.
\newblock {\em arXiv preprint arXiv:1611.07004}, 2016.

\bibitem[\protect\citeauthoryear{Jia \bgroup \em et al.\egroup }{2014}]{caffe}
Yangqing Jia, Evan Shelhamer, Jeff Donahue, Sergey Karayev, Jonathan Long, Ross
  Girshick, Sergio Guadarrama, and Trevor Darrell.
\newblock Caffe: Convolutional architecture for fast feature embedding.
\newblock {\em arXiv preprint arXiv:1408.5093}, 2014.

\bibitem[\protect\citeauthoryear{Jones \bgroup \em et al.\egroup
  }{1998}]{expectedimprovement}
Donald~R Jones, Matthias Schonlau, and William~J Welch.
\newblock Efficient global optimization of expensive black-box functions.
\newblock {\em Journal of Global optimization}, 13(4):455--492, 1998.

\bibitem[\protect\citeauthoryear{Krizhevsky \bgroup \em et al.\egroup
  }{2012}]{alexnet}
Alex Krizhevsky, Ilya Sutskever, and Geoffrey~E Hinton.
\newblock Imagenet classification with deep convolutional neural networks.
\newblock In F.~Pereira, C.~J.~C. Burges, L.~Bottou, and K.~Q. Weinberger,
  editors, {\em Advances in Neural Information Processing Systems 25}, pages
  1097--1105. Curran Associates, Inc., 2012.

\bibitem[\protect\citeauthoryear{Leong \bgroup \em et al.\egroup
  }{2003}]{leong2003correction}
FJ~WM Leong, M~Brady, and JO~McGee.
\newblock Correction of uneven illumination (vignetting) in digital microscopy
  images.
\newblock {\em Journal of clinical pathology}, 56(8):619--621, 2003.

\bibitem[\protect\citeauthoryear{Li \bgroup \em et al.\egroup
  }{2016}]{li2016fully}
Yi~Li, Haozhi Qi, Jifeng Dai, Xiangyang Ji, and Yichen Wei.
\newblock Fully convolutional instance-aware semantic segmentation.
\newblock {\em arXiv preprint arXiv:1611.07709}, 2016.

\bibitem[\protect\citeauthoryear{Lin \bgroup \em et al.\egroup }{2014}]{mscoco}
Tsung{-}Yi Lin, Michael Maire, Serge~J. Belongie, Lubomir~D. Bourdev, Ross~B.
  Girshick, James Hays, Pietro Perona, Deva Ramanan, Piotr Doll{\'{a}}r, and
  C.~Lawrence Zitnick.
\newblock Microsoft {COCO:} common objects in context.
\newblock {\em CoRR}, abs/1405.0312, 2014.

\bibitem[\protect\citeauthoryear{Mirza and
  Osindero}{2014}]{mirza2014conditional}
Mehdi Mirza and Simon Osindero.
\newblock Conditional generative adversarial nets.
\newblock {\em arXiv preprint arXiv:1411.1784}, 2014.

\bibitem[\protect\citeauthoryear{Mo{\v{c}}kus}{1975}]{bayesian}
J~Mo{\v{c}}kus.
\newblock On bayesian methods for seeking the extremum.
\newblock In {\em Optimization Techniques IFIP Technical Conference}, pages
  400--404. Springer, 1975.

\bibitem[\protect\citeauthoryear{Naha and Wang}{}]{nahaobject}
Shujon Naha and Yang Wang.
\newblock Object figure-ground segmentation using zero-shot learning.

\bibitem[\protect\citeauthoryear{Papon and Schoeler}{2015}]{papon2015semantic}
Jeremie Papon and Markus Schoeler.
\newblock Semantic pose using deep networks trained on synthetic rgb-d.
\newblock In {\em Proceedings of the IEEE International Conference on Computer
  Vision}, pages 774--782, 2015.

\bibitem[\protect\citeauthoryear{Peng \bgroup \em et al.\egroup
  }{2015}]{peng2015learning}
Xingchao Peng, Baochen Sun, Karim Ali, and Kate Saenko.
\newblock Learning deep object detectors from 3d models.
\newblock In {\em Proceedings of the IEEE International Conference on Computer
  Vision}, pages 1278--1286, 2015.

\bibitem[\protect\citeauthoryear{Ren \bgroup \em et al.\egroup
  }{2015}]{ren15fasterrcnn}
Shaoqing Ren, Kaiming He, Ross Girshick, and Jian Sun.
\newblock {Faster R-CNN}: Towards real-time object detection with region
  proposal networks.
\newblock {\em arXiv preprint arXiv:1506.01497}, 2015.

\bibitem[\protect\citeauthoryear{Rong and Yang}{2016}]{rong2016one}
Tao Rong and Ruoyu Yang.
\newblock One-shot-learning gesture segmentation and recognition using
  frame-based pdv features.
\newblock In {\em Pacific Rim Conference on Multimedia}, pages 355--365.
  Springer, 2016.

\bibitem[\protect\citeauthoryear{Ros \bgroup \em et al.\egroup
  }{2016}]{ros2016synthia}
German Ros, Laura Sellart, Joanna Materzynska, David Vazquez, and Antonio~M
  Lopez.
\newblock The synthia dataset: A large collection of synthetic images for
  semantic segmentation of urban scenes.
\newblock In {\em Proceedings of the IEEE Conference on Computer Vision and
  Pattern Recognition}, pages 3234--3243, 2016.

\bibitem[\protect\citeauthoryear{Simonyan and Zisserman}{2014}]{vgg16}
Karen Simonyan and Andrew Zisserman.
\newblock Very deep convolutional networks for large-scale image recognition.
\newblock {\em CoRR}, abs/1409.1556, 2014.

\bibitem[\protect\citeauthoryear{Sixt \bgroup \em et al.\egroup
  }{2016}]{sixt2016rendergan}
Leon Sixt, Benjamin Wild, and Tim Landgraf.
\newblock Rendergan: Generating realistic labeled data.
\newblock {\em arXiv preprint arXiv:1611.01331}, 2016.

\bibitem[\protect\citeauthoryear{Su \bgroup \em et al.\egroup
  }{2012}]{su2012crowdsourcing}
Hao Su, Jia Deng, and Li~Fei-Fei.
\newblock Crowdsourcing annotations for visual object detection.
\newblock In {\em Workshops at the Twenty-Sixth AAAI Conference on Artificial
  Intelligence}, volume~1, 2012.

\bibitem[\protect\citeauthoryear{Su \bgroup \em et al.\egroup
  }{2015}]{su2015render}
Hao Su, Charles~R Qi, Yangyan Li, and Leonidas~J Guibas.
\newblock Render for cnn: Viewpoint estimation in images using cnns trained
  with rendered 3d model views.
\newblock In {\em Proceedings of the IEEE International Conference on Computer
  Vision}, pages 2686--2694, 2015.

\bibitem[\protect\citeauthoryear{Zitnick and Doll\'ar}{2014}]{edgeBoxes}
C.~Lawrence Zitnick and Piotr Doll\'ar.
\newblock Edge boxes: Locating object proposals from edges.
\newblock In {\em ECCV}, 2014.

\end{thebibliography}

\end{document}